\pdfoutput=1

\documentclass[11pt]{article}

\usepackage[final]{acl}

\usepackage{times}
\usepackage{latexsym}
\usepackage{amssymb}
\usepackage{multirow}
\usepackage{subfigure}
\usepackage{listings}

\usepackage[T1]{fontenc}

\usepackage[utf8]{inputenc}

\usepackage{microtype}

\usepackage{inconsolata}
\usepackage{graphicx}
\usepackage{amsmath}
\usepackage[linesnumbered,ruled,vlined]{algorithm2e}
\usepackage{inconsolata}
\newcommand*\samethanks[1][\value{footnote}]{\footnotemark[#1]}
\usepackage{url}

\title{Code Needs Comments: Enhancing Code LLMs with Comment Augmentation}

\author{Demin Song\thanks{\ \  Equal contribution.}\textsuperscript{1}, Honglin Guo\samethanks\textsuperscript{1,2}, Yunhua Zhou\textsuperscript{1}, Shuhao Xing\textsuperscript{1,2}\\
 {\bf Yudong Wang\textsuperscript{1}, Zifan Song\textsuperscript{1}, Wenwei Zhang\textsuperscript{1}} \\
 {\bf Qipeng Guo\textsuperscript{1}, Hang Yan\textsuperscript{1,3}, Xipeng Qiu\textsuperscript{2}, Dahua Lin\textsuperscript{1,3}}\\
\textsuperscript{1}Shanghai AI Laboratory,  
\textsuperscript{2}School of Computer Science, Fudan University \\
\textsuperscript{3}The Chinese University of Hong Kong \\
\texttt{\{songdemin,zhouyunhua,xingshuhao.dispatch,wangyudong\}@pjlab.org.cn} \\
\texttt{\{songzifan,zhangwenwei,guoqipeng,yanhang,lindahua\}@pjlab.org.cn} \\
\texttt{\{hlguo20,xpqiu\}@fudan.edu.cn}\\
}

\begin{document}
\maketitle
\begin{abstract}
The programming skill is one crucial ability for Large Language Models (LLMs), necessitating a deep understanding of programming languages (PLs) and their correlation with natural languages (NLs). We examine the impact of pre-training data on code-focused LLMs' performance by assessing the comment density as a measure of PL-NL alignment. Given the scarcity of code-comment aligned data in pre-training corpora, we introduce a novel data augmentation method that generates comments for existing code, coupled with a data filtering strategy that filters out code data poorly correlated with natural language. We conducted experiments on three code-focused LLMs and observed consistent improvements in performance on two widely-used programming skill benchmarks. Notably, the model trained on the augmented data outperformed both the model used for generating comments and the model further trained on the data without augmentation.
\end{abstract}

\section{Introduction}
The development of Large Language Models (LLMs) has made remarkable strides across various domains, including the field of code understanding and generation. Works such as CodeGen~\cite{nijkamp2022codegen}, StarCoder~\cite{li2023starcoder}, and Code Llama~\cite{roziere2023code}  have achieved significant breakthroughs in the task of natural language to code (NL2Code). Moreover, aligning natural language descriptions with their corresponding execution code to expand code-related training corpus to further enhance the model's coding capabilities has become a research focus for scholars~\cite{yin2018learning, ahmad2021unified, wang2021codet5, neelakantan2201text, muennighoff2023octopack}. Code Llama~\cite{roziere2023code}, which is currently one of the most popular code LLMs, also mentioned that 8\% of their sample data was sourced from natural language datasets related to code.
\begin{table}
    \centering
    \resizebox{\columnwidth}{!}{
    \begin{tabular}{c|c|c|c}
        language & \#Chars of Comment & \#Chars & Comment Density \\
        \hline
        C\#    & 5.4B    & 30.8B   & 0.1764  \\
        C++        & 6.6B    & 38.0B   & 0.1753  \\
        Go         & 3.0B    & 19.6B   & 0.1553  \\
        Java       & 12B     & 66.8B   & 0.1917  \\
        JavaScript & 6.3B    & 46.9B   & 0.1352  \\
        PHP        & 5.1B    & 42.3B   & 0.1207  \\
        Python     & 9.6B    & 44.1B   & 0.2187  \\
        Ruby       & 0.9B    & 5.18B   & 0.1821  \\
        Rust       & 1.1B    & 6.44B   & 0.1641  \\
        TypeScript & 2.4B    & 20.1B   & 0.1207  \\
        \hline
        \textbf{Average}    & \textbf{5.3B}    & \textbf{32.0B}   & \textbf{0.1670}  \\
    \end{tabular}
    }
    \caption{Comment density across ten mainstream programming languages in StarCoder~\cite{li2023starcoder}. \#Chars of Comment indicates the number of non-white characters of the code comment. \#Chars is the total number of non-white characters. In fact, high quality repositories even have comment density exceeding 40\%, such as the case of mini redis\protect\footnotemark. This suggests that the existing code dataset indeed contains too few comments.
    }
    \label{comments_density}
\end{table}
\footnotetext{\url{https://github.com/tokio-rs/mini-redis}}

In fact, comments are the natural language components that are inherently related to code. \citet{guo2022unixcoder} had conducted ablation experiments to demonstrate that training models on code data with comments leads to improved ability. Moreover, the textbook and exercise data proposed by \citet{gunasekar2023textbooks}, which is considered a prior work in the field of code LLMs, can be considered a form of comment in a sense. However, generating a large amount of such data using GPT is infeasible due to cost considerations.

\begin{figure*}[!t]
    \centering
    \includegraphics[width=\linewidth]{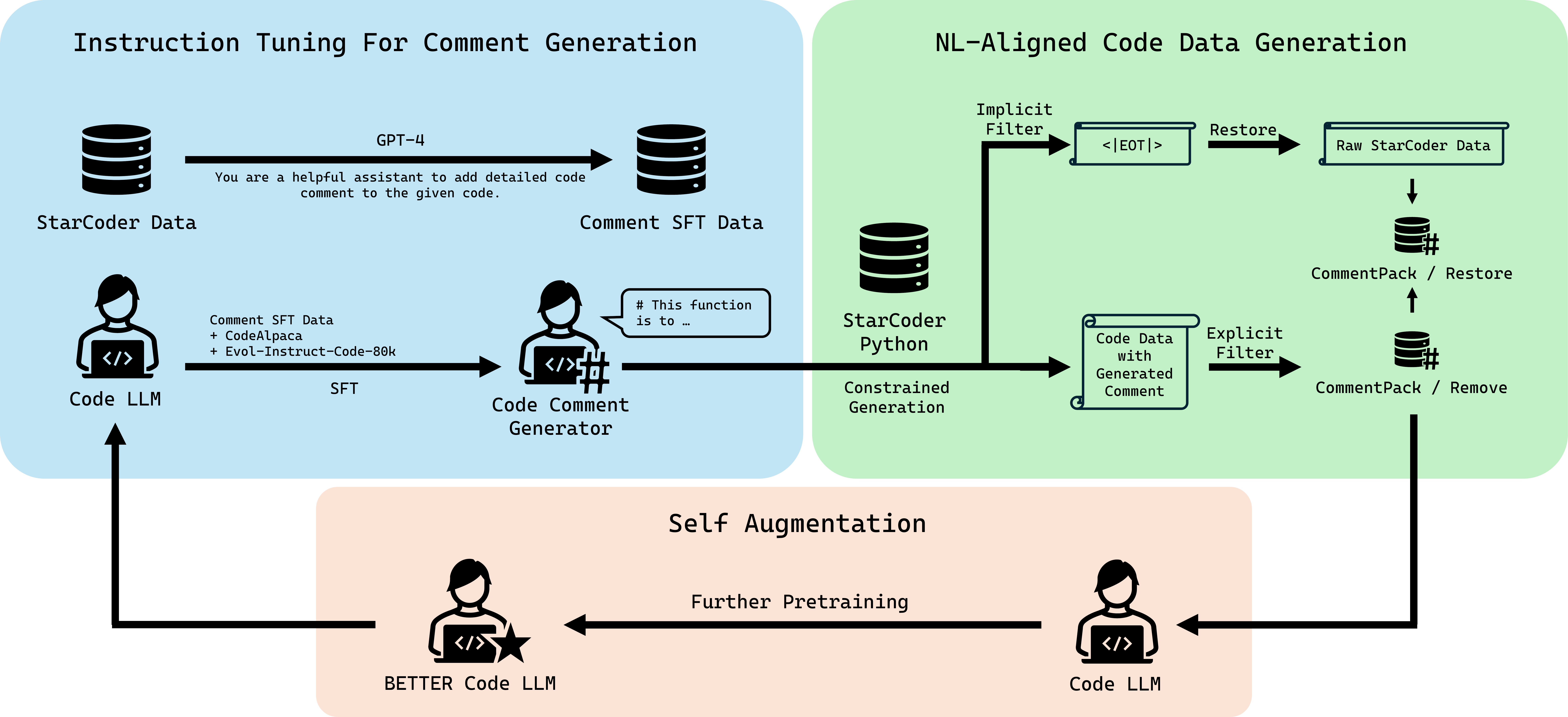}
    \centering
    \caption{Illustrates the workflow of our proposed self-augmentation method. Firstly, it enables LLMs to generate comments for code through instruction tuning. Then, LLMs generate comments for existing code. The further training is conducted on enriched code data with comments, aiming to achieve self-augmentation.}
    \label{fig:overview}
\end{figure*}

Considering that the alignment between natural language and code has not yet been relatively explored, comments serve as a representative and crucial bridge between the two. Therefore, the primary objective of this work is to explore the significance of comments. An intuitive supposition posits that an augmentation in training corpus that aligns code and natural language (comments) will invariably enhance the model's performance. To quantify this alignment, we initially delineate ``comment density'' as the ratio of the number of non-white characters in comments to the total number of non-white characters and then examine how different levels of comment density impact downstream tasks.


\begin{table*}[ht]
    \centering
    \resizebox{0.9\textwidth}{!}{
    \begin{tabular}{c|cccccc}
    \hline
        Models        & SFT                             &  Pretaining   & Natural Language   & Code & Samples & Tokens \\
        phi-1\cite{DBLP:journals/corr/abs-2306-11644}      &$\checkmark$&$\checkmark$&&$\checkmark$&-&1B\\
        WizardCoder\cite{DBLP:journals/corr/abs-2306-08568}&$\checkmark$&&&$\checkmark$&78K&-\\
        WaveCoder\cite{DBLP:journals/corr/abs-2312-14187}  &$\checkmark$&&&$\checkmark$&20K&-\\
        phi-1.5\cite{DBLP:journals/corr/abs-2309-05463}    &&$\checkmark$&$\checkmark$&&-&20B\\
        WizardLM\cite{DBLP:journals/corr/abs-2304-12244}   &$\checkmark$&&$\checkmark$&&250K&-\\
        Genie\cite{yehudai2024genie}                       &&$\checkmark$&$\checkmark$&&300K&-\\
        Self-Instruct\cite{DBLP:conf/acl/WangKMLSKH23} &
        $\checkmark$ & & $\checkmark$ & & 82K&-\\
        Ours                                               &&$\checkmark$&&$\checkmark$&6.5M&15.2B\\
    \hline
    \end{tabular}
    }
    \caption{Existing data distillation methods rely on a teacher model to acquire knowledge, and are limited by the amount of available data.}
    \label{data_distillation}
\end{table*}

As shown in Table \ref{comments_density}, existing comments in code are limited. This severely hinders our goal of improving model performance and training efficiency by increasing the amount of aligned corpus between code and natural language. Therefore, we propose a novel method aimed at generating more aligned data, which is characterized by utilizing the powerful generation capabilities of LLMs to generate comments for the original code data. To accomplish this, we require a model capable of understanding code and providing corresponding comments. From this perspective, our method can also be viewed as a form of specialized data distillation.
While, unlike traditional data distillation methods that rely on a teacher model, our approach accomplishes knowledge distillation through self-supervision. This represents the key distinction between our method and existing data distillation techniques.
Table \ref{data_distillation} provides detailed information on existing works.

To ensure that the code remains unchanged during LLMs generation and accelerate the generation process, we propose a constrained generation approach that generates data on a line-by-line basis, thereby circumventing the procedure of LLMs deleting, modifying the original code or producing new code. Considering the need to exercise caution in trusting the comments added by the model, we introduce a discriminator in this study to filter out extreme cases. The discriminator evaluates the generated comments and filters out samples that exhibit significant differences from the original code. In our experiments, we observe that utilizing LLMs for comments generation not only enhances the capabilities of the base model but also facilitates self-augmentation. The overall framework of this work is depicted  in 
Figure \ref{fig:overview} 

We highlight our contributions as follows:
\begin{itemize}
    \item We discovered that the density of comments in pre-training code significantly affects the performance of LLM models in downstream tasks, and based on this, we proposed a new data augmentation method.
    \item We introduced a new inference method for generating comments, forming an efficient self-augmentation pipeline.
    \item Our method achieved substantial improvements on 
    Llama 2, Code Llama, and InternLM2.
\end{itemize}

\section{Related Work}
\subsection{Alignment between Code and Natural Language}

\citet{yin2018learning} proposed the effective utilization of highly correlated Natural Language-Programming Language (NL-PL) pairs to enhance the capabilities of code models in tasks such as code retrieval, summarization, and generation. \citet{ahmad2021unified} employed Denoising Pre-training to establish semantic relationships between natural language and code, resulting in promising outcomes. Similarly, \citet{wang2021codet5} focused on aligning natural language and code by incorporating NL2Code and Code2NL generation tasks into the pre-training phase. \citet{neelakantan2201text} achieved superior performance over CodeBERT in the code retrieval task by employing contrastive learning to align code and natural language. \citet{muennighoff2023octopack} enhanced the code model's ability to generate code that follows natural language by utilizing commit messages.

The significance of comments as a component inherently related to code has also garnered considerable interest in research. \citet{feng2020codebert} employed the Masked Language Modeling (MLM) task on code data with comments to train a pre-trained model, yielding excellent results. \citet{wang2021syncobert}, on the other hand, utilized Contrastive Learning to align code with comments. Furthermore, \citet{guo2022unixcoder} conducted ablation experiments to demonstrate that training models on code data with comments leads to improved outcomes. In order to align natural language (NL) and code, \citet{christopoulou2022pangu} conducted a two-stage training specifically on the pairs of NL-code. This approach resulted in a significant performance improvement of approximately 70\% compared to the single-stage training. While PL-NL alignment is of paramount importance, it is challenging to obtain naturally aligned data at the scale required for pre-training purposes.T herefore, we employ LLMs to generate corresponding natural language expressions based on the existing code.

\subsection{Data Augmentation in the Field of Code}

Code augmentation techniques can be categorized into Rule-based Techniques and Model-based Techniques. Rule-based methods often involve techniques such as replacing variable names, renaming method names, and inserting dead code to transform code snippets. Some code transformations also consider deeper structural information, such as control-flow graphs (CGFs) and use-define chains (UDGs)~\cite{quiring2019misleading}. Model-based Techniques commonly utilize pre-trained models to replace non-keywords in the original data~\cite{song2022not}. Another approach employed is similar to Back-Translation, where code translation tasks are augmented by translating between two programming languages using natural language as an intermediate language ~\cite{sennrich2015improving}.

In addition, there are also several methods based on Example Interpolation Techniques. For instance, \citet{dong2022enhancing} merges rule-based techniques for source code models with mixup to blend the representations of the original code snippet and its transformed counterpart. \citet{li2022exploring} introduces two novel interpolation techniques, namely Binary Interpolation and Linear Extrapolation, for source code models. Diverging from the aforementioned approach, we present a novel methodology as the pioneering endeavor to enhance comments by leveraging existing code.

\subsection{Data Distillation in the Field of LLMs}

In this work, our approach of data augmentation through the utilization of  LLMs can be regarded as a form of data distillation. Such tasks typically rely on two processes: generation and filtering. Unnatural Instructions and Self-Instruct~\cite{DBLP:conf/acl/HonovichSLS23, DBLP:conf/acl/WangKMLSKH23} have employed this method in the creation of an instruction dataset. While following the aforementioned two steps, WizardLM and WizardCoder~\cite{DBLP:journals/corr/abs-2304-12244, DBLP:journals/corr/abs-2306-08568} utilized an Instruction Evolver to generate more diverse data. In fact, as the competency of the Teacher model has advanced, numerous studies have gradually phased out  the step of using a discriminator to filter data~\cite{DBLP:journals/corr/abs-2306-11644, DBLP:journals/corr/abs-2309-05463}.

However, the data generated by these methods all originates from the Teacher model, which often limits them to the knowledge of the Teacher. To mitigate this limitation, GENIE~\cite{yehudai2024genie} proposes generating task-specific examples from the content. Similarly, in WaveCode~\cite{DBLP:journals/corr/abs-2312-14187}, the code generation task involves generating instructions from code. Taking a step further, our method completely liberates itself from the constraints of a teacher model, enabling highly efficient generation of large-scale pre-training data.

\section{Method}

Indeed, generating comments for existing code by using LLMs is not a simple task for us with two principal challenges. Firstly, LLMs often struggle to effectively follow the ``add comments'' instruction, resulting in code loss or insufficient comment additions, especially for longer code files. Secondly, generating comments for large-scale pre-training code data can be computationally expensive, leading to significant training costs for the entire model. Appendix \ref{sec:bad_case_1} is a bad case where LLMs fail to follow the instruction of ``add comments''. 

\subsection{Instruction Tuning for Comment Generation}

In order to endow LLMs with the capacity to rigorously follow ``add comments'' instructions, we deliberately constructed an Instruction dataset for fine-tuning LLMs.

\paragraph{Instruction Dataset}
\begin{table}
    \centering
    \resizebox{\columnwidth}{!}{
    \begin{tabular}{c|ccccc}
    \hline
        language        & c-sharp & cpp   & go   & java & javascript \\
        Instruct Num    &  447    & 364   & 425  & 435  & 458        \\
        \hline
        language        & python  & php   & ruby & rust & typescript\\
        Instruct Num    & 495     & 449   & 466  & 391  & 462  \\
    \hline
    \end{tabular}
    }
    \caption{We constructed over 4000 instruction data from a total of 10 mainstream code of StarCoder~\citep{li2023starcoder}.}
    \label{instruct_data}
\end{table}
In this work, we selected over 4000 samples from the 10 distinguished programming languages discussed in StarCoder Datasets~\cite{li2023starcoder}. These samples were then augmented with corresponding comments using the GPT-4 model~\citep{openai2023gpt}, resulting in the creation of an extensive instruction dataset. Following a meticulous manual screening process, we refined the dataset, retaining a total of 4394 high-quality instruction data instances. Then, we convert the prompt and code into Markdown format. Please find the sample of our instruction data from Appendix \ref{sec:sft_data}

To mitigate the risk of the model overfitting to the specific characteristics of the instruction data, we incorporated additional datasets: CodeAlpaca~\cite{codealpaca} and Evol-Instruct-Code-80k~\cite{luo2023wizardcoder}. To ensure the uniqueness of our instructions, we meticulously removed any instruction data with comments that overlapped with the CodeAlpaca and Evol-Instruct-Code-80k datasets. After creating instruction data, we use it to finetune our base model: CodeLlama-7b~\cite{roziere2023code} and obtain a code comments generator.

For a comprehensive overview of the language distribution within our instruction dataset for comment generation, please refer to Table \ref{instruct_data}

\paragraph{Implicit Filter}
\begin{figure}[t]
    \includegraphics[width=1.0\linewidth]{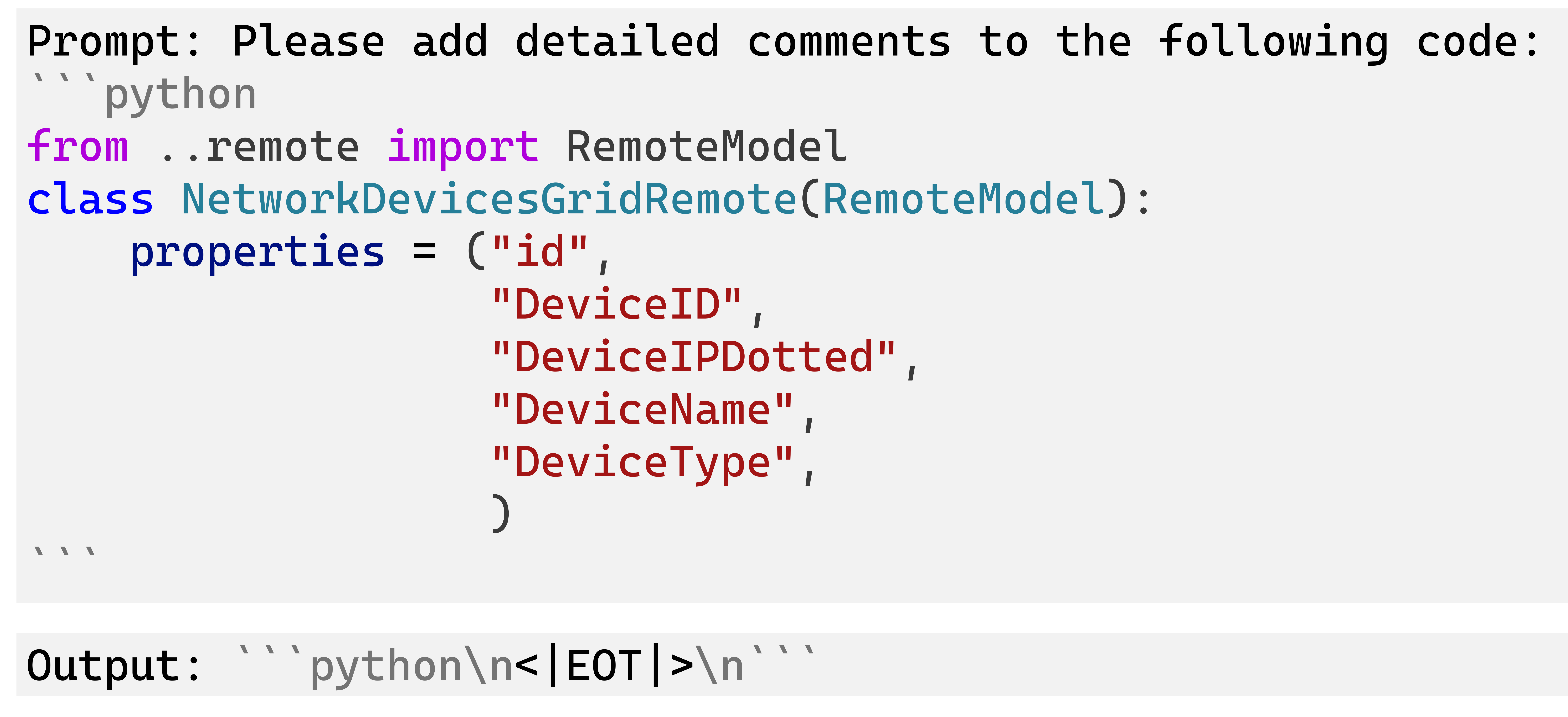}
    \caption{If the LLM discovers code with low training value, it will output <|EOT|> to implement an implicit filtering mechanism.}
    \label{fig:implicit_filter}
\end{figure}
Although the StarCoder~\cite{li2023starcoder} dataset underwent certain filtering processes, there are still some data instances that lack training value (e.g., containing only module imports, version specifications, or very simple class definitions). To counteract this predicament, we incorporated particular samples within the instruction datasets, wherein the output was designated as ``<|EOT|>'' to signify that the model does not deem the input code is worth adding comments. This strategy is designed with the objective of endowing the model with the capacity to recognize high-quality code data throughout the process of comments generation. Figure \ref{fig:implicit_filter} provides an example of such a sample.

\subsection{NL-Aligned Code Data Generation}

To ensure the preservation of the original code during the comments generation process and to facilitate a degree of acceleration, we introduce a novel method of constrained generation. Indeed, preservation of the original code is crucial to avoid the model generating illusory, repetitive code. Further details and information regarding this aspect can be found in the Appendix \ref{sec:bad_case_2}
\paragraph{Constrained Generation}

\begin{figure*}[!t]
    \centering
    \includegraphics[width=1.0\linewidth]{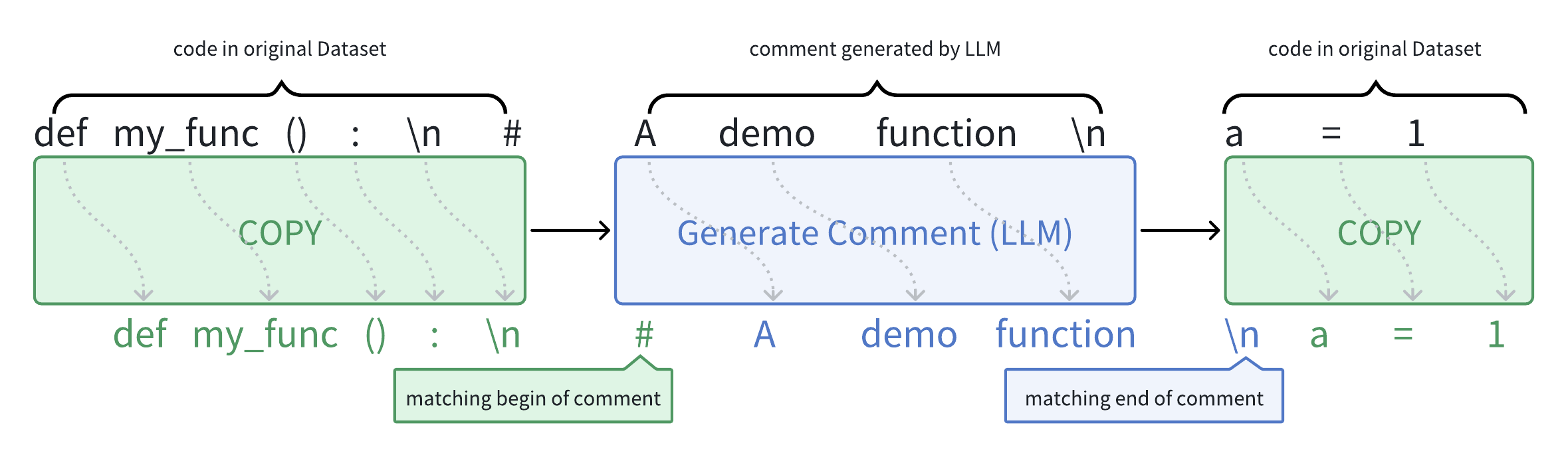}
    \centering
    \caption{Illustration of the constrained generation algorithm. During the generation process, the code will be directly copied into the output until it encounters the marker indicating the beginning of a comment (\#, ''' or """ for Python). The commented portion is generated by the code comment generator until the end of the comment (\textbackslash n, ''' or """, correspondingly).}
    \label{fig:constrained_generation}
\end{figure*}

 In the task of generating comments for existing code, there is a notable characteristic in the LLM's decoding stage: the generated content of the model can be easily separated into comments and code on a line-by-line basis. Since the code is precisely the input given to the model, we can directly skip the process of generating code by the model. 

More formally, let $C=\{C_{i}\}$ represent the code data for which comments are to be generated, where $C_{i}$ denotes the $i$-th line of the code. Let $x=\{\text{prompt}, C\}$ be the input sequence, and $y^l_{t}$ be the $t$-th token generated by the LLM in the $l$-th line. It is worth noting that this generation process is performed on a line-by-line basis.




\begin{equation}
    y^l_{t} \sim \begin{cases}
        P(y|x, y^{<l}, y^{l}_{<t}) & \text{$y^l_{<t}$ is comment}, \\
        C_{j} & \text{$y^l_{<t}$ is code}.
    \end{cases}  
\end{equation}



In fact, during the process of generating each line of data of LLMs, it is possible to determine whether a particular line is code or not by using regular expressions with just a few initial tokens.

Please refer to Algorithm \ref{algorithm_constrained_generation} for the pseudo code and Figure \ref{fig:constrained_generation} for an illustration of our method.

\begin{algorithm}[t]
    \small
    \SetKwInOut{Input}{Input}
    \SetKwInOut{Output}{Output}
    \SetKwFunction{LLM}{LLM}
    \SetKwFunction{gencode}{gen\_code}
    \SetKwFunction{APPEND}{APPEND}
    \SetKwFunction{EXTEND}{EXTEND}
    \SetKwFunction{POP}{POP}
    \SetKwFunction{stop}{stop}
    
    \Input{$x$, $C=\{C_1, \dots, C_n\}$}
    \Output{$y$}
    
    \BlankLine
    $y \leftarrow []$\;
    
    \While{true}{
        $o \leftarrow \LLM(x, y)$\;
        
        \uIf {\textnormal{not} \gencode($y$, $o$)}{
            \APPEND($y$, $o$)\;
        }
        \Else{
            \EXTEND($y$, \POP($C$))\;
        }
        
        \uIf{\stop($y$)}{
            \textbf{break}\;
        }
    }

    \caption{Constrained Generation}
    \label{algorithm_constrained_generation}
\end{algorithm}

\paragraph{Explicit Filter}
To exclude exceedingly poor instances in the comments generated by LLMs and ensure the quality of generated comments, 
we introduced two additional filtering rule:
\begin{itemize}
    \item Excluding code data generated by LLMs that does not adhere to the markdown format.
    \item Excluding code data generated by LLMs where the discrepancy in length between the generated code and the original code exceeds 100\%.
\end{itemize}
\subsection{Self Augmentation}
Upon executing the aforementioned two processes, we will acquire a high-quality code dataset with extensive comments. We can then proceed to conduct additional training to augment the capabilities of our base model, resulting in a better code LLM. This process engenders a self-augmentation feedback loop. Subsequently, the better LLm model will serve as the base code LLm for the next iteration of self-augmentation, to be performed repeatedly. The overall process of our approach is illustrated in Figure \ref{fig:overview}.

\section{Experiments}
We initially lay the foundation with empirical evidence on the Llama 2 model~\cite{touvron2023llama}, illustrating that the fortification of alignment between code and natural language—particularly through the amplification of comment density—profoundly influences downstream tasks. Subsequently, we apply our proposed methodology to the Code Llama model~\cite{DBLP:journals/corr/abs-2308-12950}, underscoring its capacity not merely to bolster weak baselines such as Llama 2, but also to achieve self-augmentation on models like Code Llama, distinguished by their exceptional performance in code generation tasks. Moreover, we have substantiated through the InternLM2~\cite{2023internlm} which is the most recent state-of-the-art LLm in the field. that the PL-NL alignment data, generated by CodeLLama, retains its efficacy for other models. All models were validated on the HumanEval~\cite{DBLP:journals/corr/abs-2110-14168} and MBPP~\cite{DBLP:journals/corr/abs-2108-07732} datasets.
\subsection{Dataset}

As an initial step, we elected to utilize the Python data from StarCoder~\cite{li2023starcoder} as our experimental validation dataset, henceforth referred to as \textbf{SP} (StarCoder Python) to circumvent any potential confusion. Leveraging the instruct data formulated in the preceding section, we enacted instruct tuning on the CodeLlama-7b model, thereby equipping it with the capability to generate comments for code. This model was subsequently employed to append comments to the SP dataset.

Owing to the existence of code data in StarCoder, characterized by an excessive number of tokens, the procedure of incorporating comments frequently surpasses the model's maximum sequence length. Consequently, we opted to exclude this subset of data from the comment addition process, preserving it for subsequent datasets.

Within our approach, we integrated both implicit and explicit filters to ensure the integrity of the code data and the generated comments. As a result, a considerable proportion of data was unable to pass through the implicit filter (model outputting <|EOT|>) or the explicit filter during the comment generation process. We adopted two distinct strategies to address this situation:

\begin{itemize}
    \item Discarding the data that failed to traverse the implicit or explicit filter, culminating in a superior-quality dataset labeled \textbf{CommentPack / Remove} (CP/Remove, remove <|EOT|> samples in comment-packed python data).
    \item Substituting the model's output with the original code data for instances that were unable to pass through either filter, leading to a lower-quality dataset (maintaining the same scale as the original dataset), designated as the \textbf{CommentPack / Restore} (CP/Restore, substitute raw StarCoder data for <|EOT|> samples in comment-packed python dataset) dataset.
\end{itemize}

\begin{table}
    \centering
    \resizebox{\columnwidth}{!}{
    \begin{tabular}{c|cccc}
    \hline
        Dataset & \#Samples & Comment Density (\%) & \#Tokens \\
        StarCoder Python  & 12.8M & 21.87 & 20.8B \\ 
        StarCoder Python / Remove  & 6.54M & 23.08 & 13.1B \\ 
        StarCoder Python / Absent & 12.8M &   0.0 & 16.7B \\ 
        CommentPack / Restore & 12.8M & 32.59 & 21.5B \\ 
        CommentPack / Remove & 6.54M & 38.23 & 15.2B \\ 
    \hline
    \end{tabular}
    }
    \caption{Number of samples, comment density and number of tokens of the corresponding code datasets.}
    \label{dataset}
\end{table}


\begin{table*}[ht]
\centering
    \resizebox{\textwidth}{!}{
\begin{tabular}{llllllll}
\hline
\multirow{2}{*}{MODEL} & \multicolumn{1}{l|}{\multirow{2}{*}{DATA}}      & \multicolumn{3}{c|}{HumanEval}                    & \multicolumn{3}{c}{MBPP}      \\
                       & \multicolumn{1}{l|}{}                           & pass@1 & pass@5 & \multicolumn{1}{l|}{pass@10} & pass@1 & pass@5 & passs@10 \\ \hline

Llama2-7b    & \multicolumn{1}{l|}{-} & 12.25 & 19.75 & \multicolumn{1}{l|}{23.73}  &  20.81 & 29.10  & 37.75 \\
Llama2-7b    & \multicolumn{1}{l|}{SP/Absent} & 16.46 & 27.87 & \multicolumn{1}{l|}{34.22}  & 19.00   & 40.10  & 48.16 \\
Llama2-7b    & \multicolumn{1}{l|}{SP}  & 17.07 & 31.09 & \multicolumn{1}{l|}{\textbf{39.06}}  & 20.40 & \textbf{52.45} & \textbf{50.90}  \\
Llama2-7b    & \multicolumn{1}{l|}{CP/Restore}  & \textbf{23.17}  & \textbf{31.79} & \multicolumn{1}{l|}{38.84} & \textbf{29.20}  &   41.20    &  49.34      \\
CodeLlama-7b & \multicolumn{1}{l|}{-}              & 31.10       &    45.75    & \multicolumn{1}{l|}{56.81}        &   42.80     &  56.50      &  64.82        \\
CodeLlama-7b & \multicolumn{1}{l|}{SP}& 32.32 & 43.70 & \multicolumn{1}{l|}{53.41} & \textbf{45.00}   & \textbf{58.03}  & \textbf{65.41}  \\
CodeLlama-7b &  \multicolumn{1}{l|}{SP/Remove}& 33.54 & 46.87 &     \multicolumn{1}{l|}{57.33} & 44.80   & 57.68 & 65.23 \\
CodeLlama-7b & \multicolumn{1}{l|}{CP/Restore} & 32.32  & 47.81 & \multicolumn{1}{l|}{57.27}        & 44.20   & 57.10 &  64.97 \\
CodeLlama-7b & \multicolumn{1}{l|}{CP/Remove}  & \textbf{39.02}  & \textbf{51.89} & \multicolumn{1}{l|}{\textbf{61.50}} & 43.00 & 56.70 & 64.99 \\
InternLM2-7b-base & \multicolumn{1}{l|}{-} & 32.32 & 49.64 & \multicolumn{1}{l|}{60.13} & 41.40 & 54.06 & 62.23 \\
InternLM2-7b-base       & \multicolumn{1}{l|}{SP}    & 35.98      & 49.82       & \multicolumn{1}{l|}{59.57}        & 43.00       & 56.24   & 64.18       \\
InternLM2-7b-base  & \multicolumn{1}{l|}{CP/Remove}  &  \textbf{40.20}     &   \textbf{50.90}      & \multicolumn{1}{l|}{\textbf{60.78}}         & \textbf{43.00}        &     \textbf{56.87}    &    \textbf{64.99}       \\ 
InternLM2-7b       & \multicolumn{1}{l|}{-}  & 43.29 & 56.31 & \multicolumn{1}{l|}{67.64} & 44.00 & 57.72 & 63.10 \\
InternLM2-7b       & \multicolumn{1}{l|}{SP} & 42.70 & \textbf{59.67} & \multicolumn{1}{l|}{\textbf{70.72}} & 42.60 & 61.61 & 67.15\\
InternLM2-7b       & \multicolumn{1}{l|}{CP/Remove} & \textbf{49.39} & 58.04 & \multicolumn{1}{l|}{68.27} & \textbf{47.80} & \textbf{64.89} & \textbf{71.12} \\
\hline
\end{tabular}
    }
    \caption{Experiment results of further pre-training. 
    "-" indicates the origin model without tuning. Almost all of the base models achieved leading performance on dataset SC/Remove, especially in the results of Pass@1.}
    \label{codellama_result}
\end{table*}
Moreover, to streamline comparisons with the CP/Remove dataset, we gathered the corresponding original data for these instances, thereby constructing the \textbf{StarCoder Python / Remove} (SP/Remove, remove <|EOT|> samples in original python dataset of StarCoder) dataset. 

In addition, to validate the importance of comments in the code dataset, we utilized regular expressions to eliminate all comments from the SPO dataset, thus creating a pure code dataset. This dataset solely consists of code samples without any accompanying comments, named \textbf{StarCoder Python / Absent} (SP/Absent, means the absence of comments in the python dataset of StarCoder)
Table \ref{dataset} provides a detailed overview of the datasets mentioned.

\subsection{Training Details}
\paragraph{Further Training}
 
Our optimizer is AdamW \cite{DBLP:conf/iclr/LoshchilovH19} with $\beta_{1}$ and $\beta_{2}$ value of 0.9 and 0.95. We use a cosine scheduler with 250 warm-up steps, and set the final learning rate to be 1/10 of the peak learning rate. We use a batch size of 4M tokens which are presented as sequences of 4,096 tokens for Llama 2, 16384 tokens for Code Llama and InternLM 2. 40B tokens in total. We set the initial learning rate to $1e^{-5}$ for Llama 2, $3e^{-6}$ for Code Llama and InternLM2. 

\paragraph{Instruction Training}
To further assess the performance of our model, we conducted instruction tuning using the dataset proposed by AlchemistCoder\cite{anon}. The training was performed with a batch size of 512K tokens, organized as sequences of 8192 tokens. We employed a learning rate of $1e^{-5}$ and trained the model for 2 epochs on a cluster consisting of 32 NVIDIA A100-80GB GPUs.

\subsection{Data Distillation}

Table \ref{codellama_result} shows the experimental results conducted on the Llama2-7b model. The results clearly demonstrate that as the comment density increases (with a comment density of 0 for ``SP/Absent'' and a density of 38.23\% for ``CP/Remove''), the model's performance exhibits significant improvements  transitioning from 16.46 to 23.17 on  HumanEval dataset, 19.00 to 29.20 on MBPP dataset. 
\begin{figure*}[!t]
    \centering
    \subfigure[Result of further pre-training on Llama 2 7B, CD means Comment Density]{
        \includegraphics[width=.44\linewidth]{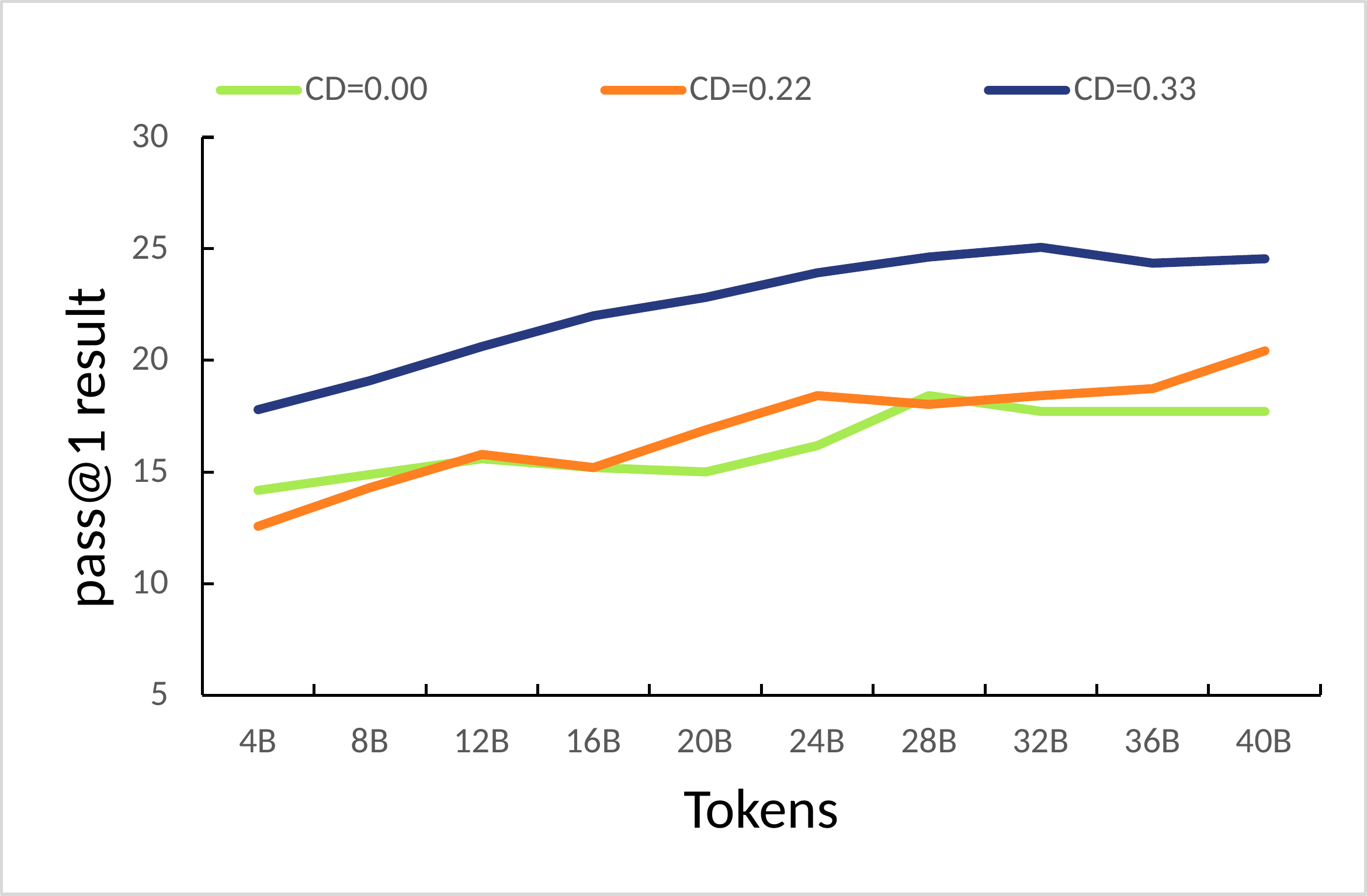}
        \label{fig:llama_token}
    }
    \subfigure[Result of further pre-training on Code Llama 7B]{
        \includegraphics[width=.44\linewidth]{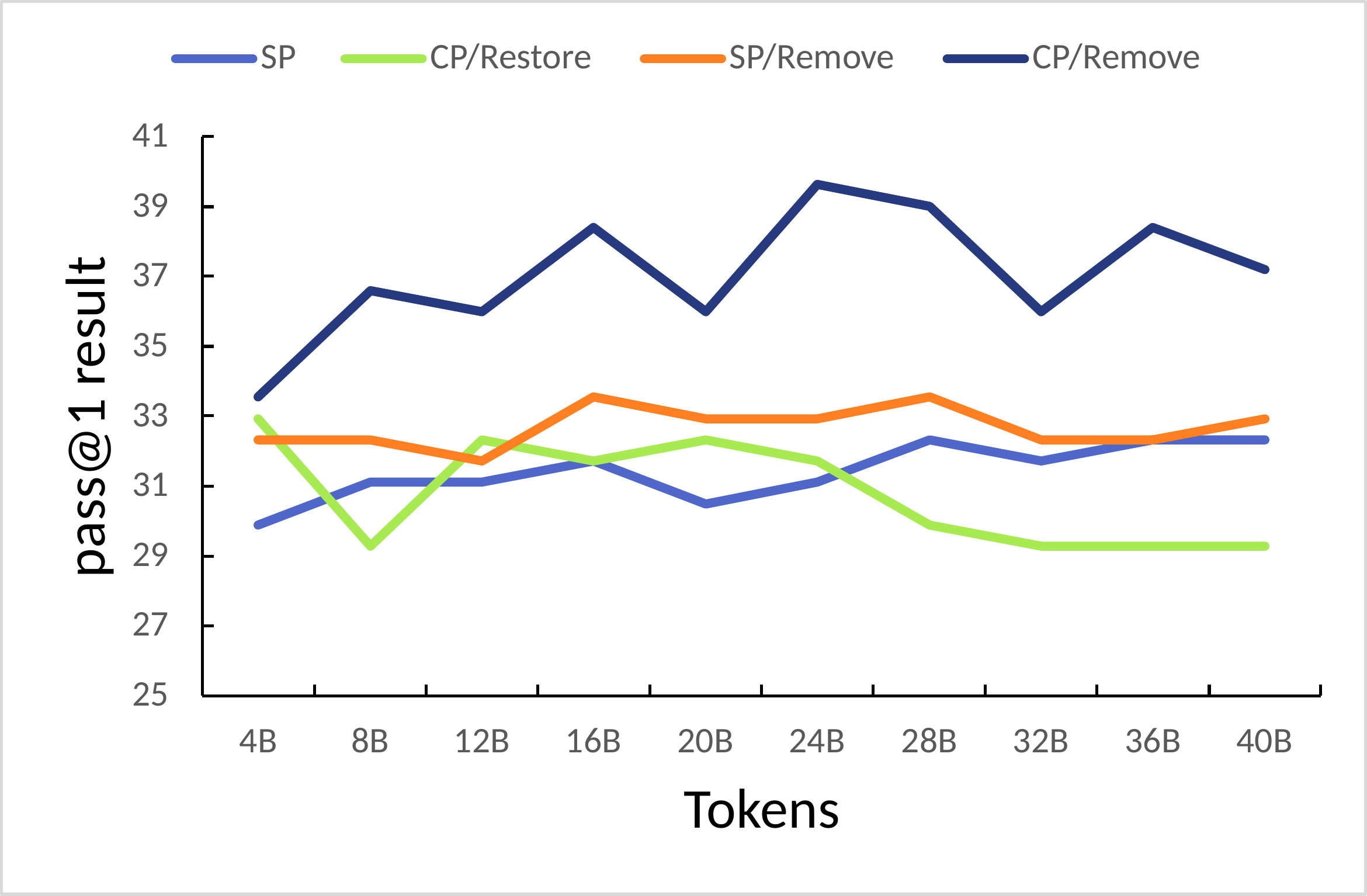}
        \label{fig:codellama_token}
    }
    \caption{HumanEval performance variation with respect to the number of training tokens.}
    \label{fig:token}
\end{figure*}

From Figure \ref{fig:llama_token}, it is clear that when training with the same number of tokens, data with a higher comment ratio achieves better results in downstream tasks. This result indicates that, under the same amount of data, a higher comment density makes it easier to learn the code, improves the alignment between natural language and code, and is more beneficial for code generation-oriented downstream tasks

\subsection{Self-Augmentation}

Firstly, Table \ref{codellama_result} provides a comprehensive overview of the results obtained from Further Training of Code Llama on the SP and CP/Restore datasets. The analysis reveals that merely replacing the filtered data, removed by explicit and implicit filters, with the original data does not significantly improve the model's performance on downstream tasks. However, when the filtered data is completely removed (as observed in Code Llama's results on SP and SP/Remove), a certain degree of improvement can be observed on the HumanEval evaluation set. Although this improvement may not be substantial, it still underscores the necessity of the filters. Similar conclusions can be drawn from the comparison of Code Llama's further training results on CP/Restore and CP/Remove datasets.

\begin{table}
    \centering
    \resizebox{\columnwidth}{!}{
    \begin{tabular}{cc|cccc}
    \hline
        MODEL          & DATA & HumanEval   & MBPP  \\
        \hline
        CodeLlama-7b   & -    & 63.40    & 53.20 \\
        CodeLlama-7b   & SP   & \textbf{66.46}    & 55.80 \\
        Instruct Num   & CP/Remove & 65.85 & \textbf{58.60} \\
    \hline
    \end{tabular}
    }
    \caption{Experiment Pass@1 result in HumanEval and MBPP of Instruction Fine-tuning.
    "-" indicates the origin model without tuning.}
    \label{sft_result_pass_1}
\end{table}
For the same filtered data, the addition of more comprehensive comments leads to significant performance gains on HumanEval after further training (as evident from Code Llama's results on CP/Remove and CP/Restore). However, it should be acknowledged that the structure of MBPP's data and the way we incorporate data into the code differ significantly, and we did not achieve substantial improvements during the further training phase on MBPP. Nevertheless, we discovered that this does not imply a lack of substantial performance enhancement for the model. In fact, as show in Table \ref{sft_result_pass_1}, when Code Llama undergoes instruction tuning after further pre-training on SP and CP/Remove datasets, it further enhances the model's adaptability to the MBPP dataset, resulting in a noteworthy improvement of 5.4\% pass@1 on CP/Remove. Please refer to the Appendix \ref{sec:appendix_instruction_fine_tuning} for the results of Pass@5 and Pass@10.

Furthermore, the comment generated by our approach on Code Llama remain effective for other models as well (as demonstrated by the comparison with further training results on SP and CP/Remove of InternLM2, where Code Llama's comments yield a significant improvement of 6\% pass@1 on HumanEval for the InternLM2-7b-base model, 6.6\% pass@1 on HUmanEval, 5.2\% pass@1 on MBPP for the InternLM2-7b model).

Lastly, Figure \ref{fig:codellama_token} demonstrates that the data quality of SP/Remove surpasses that of SP. Furthermore, after incorporating comments into SP/Remove (CP/Remove), there is a significant qualitative improvement in the dataset's quality. This leap in data quality can be observed if we acknowledge the close correlation between data quality and downstream tasks, under the assumption that the base model remains consistent.

\subsection{Constrained Generation}
\begin{figure}
    \hspace{-1.5cm}
    \includegraphics[width=1.3\linewidth]{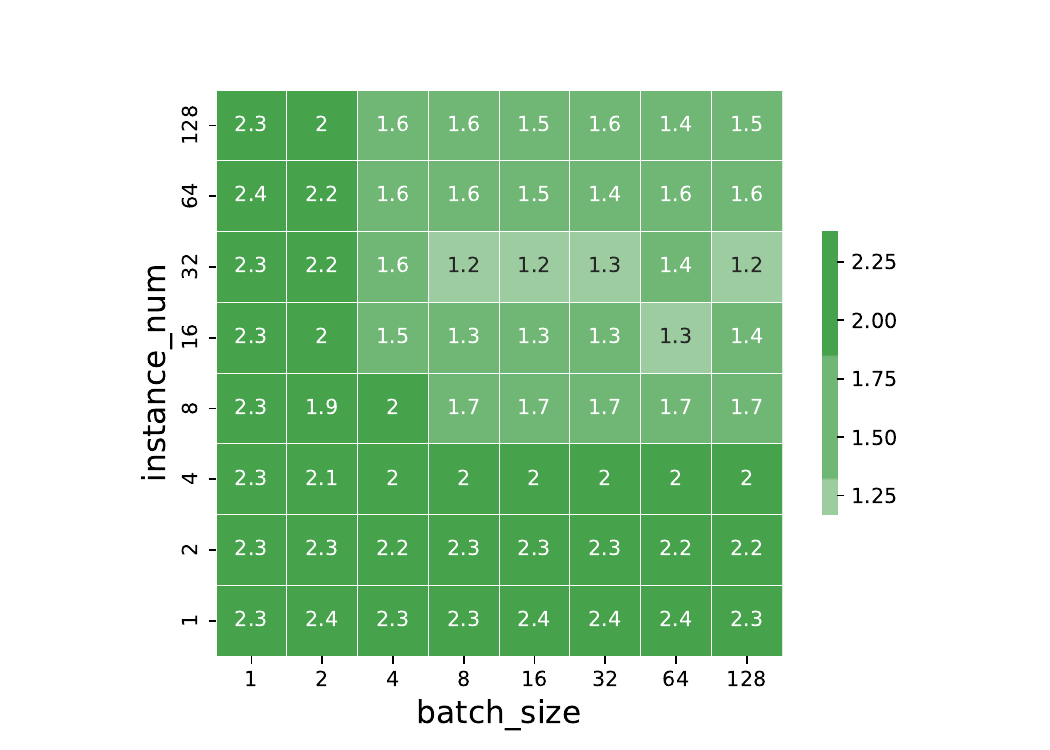}
    \caption{Heat map of speedup ratio across different combinations of instance numbers and batch sizes.}
    \label{fig:heatmap}
\end{figure}

We have implemented the Constraint Generation method on LMDeploy\footnote{\url{https://github.com/InternLM/lmdeploy}} and demonstrated its effectiveness in accelerating decoding under different experimental. Despite LMDeploy already incorporating various acceleration techniques such as page attention, our method exhibits notable speed improvements.

As evident from Figure 5, the results indicate that our method achieves the most significant acceleration when the batch size and instance number are relatively small. Even when the GPU is operating at maximum capacity (e.g., batch\_size=128, instance\_num=128), our method still provides a certain degree of speed enhancement.


\section{Conclusion}
In this paper, we propose a novel method of code data augmentation that generates comments for existing code. We validate its effectiveness on three different LLMs. This signifies a novel paradigm shift towards self-augmentation for code LLMs, thereby illuminating the latent potential for LLMs to self-evolve and enhance.

\section{Limitation}
In this paper, although we have successfully eliminated the reliance on data distillation with a teacher model, it is important to note that performing data augmentation on the pre-training dataset still incurs considerable GPU overhead. Additionally, using "<|EOT|>" as the model's output in the implicit filter stage may not align well with the behavioral patterns typically exhibited by a language model. It might be more beneficial to consider using natural language instead. Furthermore, during the next iteration of self-augmentation, we observed only marginal improvements, which is why these results were not reported in the main experiments. Further exploration and investigation are needed in this regard.



\bibliography{custom}

\appendix

\section{Bad Cases of Comment Generation by LLMs}
\label{sec:bad_case_1}

\definecolor{codegreen}{rgb}{0,0.6,0}
\definecolor{codegray}{rgb}{0.5,0.5,0.5}
\definecolor{codepurple}{rgb}{0.58,0,0.82}
\definecolor{backcolour}{rgb}{0.95,0.95,0.92}

\lstdefinestyle{pythonstyle}{
    language=Python,
    basicstyle=\small\ttfamily,
    keywordstyle=\color{codepurple},
    stringstyle=\color{codegreen},
    commentstyle=\color{codegray}\itshape,
    backgroundcolor=\color{backcolour},
    frame=tb,
    numbers=left,
    numberstyle=\tiny\color{codegray},
    breaklines=true,
    showstringspaces=false,
    tabsize=4,
    captionpos=b,
    morekeywords={True, False},
    escapeinside={(*@}{@*)}
}

\lstset{style=pythonstyle}
\begin{lstlisting}[style=pythonstyle, caption={a bad case where LLMs fail to follow the instruction of ``add comments''.}, label={code:bad_case_1}]
Prompt: Please add detailed comments to the following code
```python
# https://www.hackerrank.com/challenges/save-the-prisoner

# this is too slow


def save(n, sweets, s):
    sweets -= 1
    while sweets:  # while we have candy
        if s == n:  # if we reach the end, reset
            s = 0
        else:
            s += 1  # increase the id
        sweets -= 1
    return s


t = int(input().strip())
for a0 in range(t):
    n, m, s = input().strip().split(' ')
    n, m, s = [int(n), int(m), int(s)]

    print(save(n, m, s))

```
Model Output:
```python
# This code is designed to solve the "Save the Prisoner" problem on HackerRank.
# The problem is as follows:

# There are N prisoners standing in a circle waiting to be executed. The executions occur around a circular table, and each execution takes exactly one minute. The executions are carried out starting from the prisoner at position 1 and moving clockwise until the last prisoner is executed. The executions are carried out such that each prisoner is guaranteed to be executed.

# Given N, the total number of prisoners, M, the number of executions per hour, and S, the position of the prisoner to be executed next, find the position of the prisoner to be executed in the Nth hour.

# The provided code is an implementation of a solution to this problem. It uses a while loop to simulate the executions of the prisoners, decrementing the number of remaining sweets with each execution.

# The code first decrements the number of remaining sweets by 1. Then, it checks if there are any remaining sweets. If there are, it continues the loop.

# If the prisoner to be executed next (S) is equal to N, it means that the circle has returned to the beginning, so it resets S to 0. Otherwise, it increments S by 1 to move to the next position.

# Finally, it returns the value of S, which represents the position of the prisoner to be executed in the Nth hour.

# This code is too slow because it does not take into account the fact that the executions occur around a circular table, which means that the order of executions may not be in a strict linear sequence.
# A more efficient solution would be to consider the fact that each execution takes exactly one minute, and the circle returns to the beginning after M executions per hour.
# This would allow the code to predict the position of the prisoner to be executed in each hour, and thus find the position of the prisoner to be executed in the Nth hour.
```
\end{lstlisting}

\section{A Sample of Instuctions Data}
\label{sec:sft_data}
\begin{lstlisting}[style=pythonstyle, caption={a sample of our instuctions data.}, label={code:bad_case_2}]
Prompt: Please add detailed comments to the following code 
```python
import sys
import os
import glob

if len(sys.argv) != 2:
  print("Error: wrong format.\nUsage: python find_class.py [class_name]")
  sys.exit(0)

searching_class_name = sys.argv[1]

def find_class(class_name):
  file_list = glob.glob('*.txt')
  file_list.sort()
  # iterate through the text files
  file_found = False
  for txt_file in file_list:
    # open txt file lines to a list
    with open(txt_file) as f:
      content = f.readlines()
    # remove whitespace characters like `\n` at the end of each line
    content = [x.strip() for x in content]
    # go through each line of eache file
    for line in content:
      class_name = line.split()[0]
      if class_name == searching_class_name:
        print(" " + txt_file)
        file_found = True
        break
  if not file_found:
    print(" No file found with that class")

print("Ground-Truth folder:")
os.chdir("../ground-truth")
find_class(searching_class_name)
print("\nPredicted folder:")
os.chdir("../predicted")
find_class(searching_class_name)

```
Model Output:
```python
# import necessary libraries
import sys
import os
import glob

# check if the command line arguments are valid
if len(sys.argv) != 2:
    print("Error: wrong format.\nUsage: python find_class.py [class_name]")
    sys.exit(0)

# get the class name from the command line arguments
searching_class_name = sys.argv[1]


# function to find the class in the text files
def find_class(class_name):
    # get all text files in the current directory
    file_list = glob.glob("*.txt")
    file_list.sort()
    # flag to check if the class is found in any file
    file_found = False
    # iterate through the text files
    for txt_file in file_list:
        # open txt file lines to a list
        with open(txt_file) as f:
            content = f.readlines()
        # remove whitespace characters like `\n` at the end of each line
        content = [x.strip() for x in content]
        # go through each line of each file
        for line in content:
            class_name = line.split()[0]
            # if the class name matches the input class name
            if class_name == searching_class_name:
                print(" " + txt_file)
                file_found = True
                break
    # if the class is not found in any file
    if not file_found:
        print(" No file found with that class")


# find the class in the ground-truth folder
print("Ground-Truth folder:")
os.chdir("../ground-truth")
find_class(searching_class_name)

# find the class in the predicted folder
print("\nPredicted folder:")
os.chdir("../predicted")
find_class(searching_class_name)

```
\end{lstlisting}

\section{Bad Cases of Original Generation}
\label{sec:bad_case_2}
\begin{lstlisting}[style=pythonstyle, caption={a bad case where an LLM generates an illusion using the original generation approach.}, label={code:bad_case_2}]
Prompt: Please add detailed comments to the following code 
```python
import functools
import logging


@functools.lru_cache
def component_logger(name: str):
    return logging.getLogger(name)

```
Model Output:
```python
import functools
import logging


# Decorator to cache the results of the function, using the function arguments as the cache key
# This will make the function faster by returning the cached result when the same inputs are provided
@functools.lru_cache
def component_logger(name: str):
    # This function is responsible for creating a logger object for a given component
    # The logger object is used to log messages for a specific component

    # Get the logger for the given component name
    # The logger is a singleton object that can be used to log messages for a specific component
    # The 'name' parameter is used to identify the component for which the logger is being created
    logger = logging.getLogger(name)

    # Set the log level to INFO
    # This is the default log level in Python logging module, which means that only log messages of level INFO or higher will be displayed
    logger.setLevel(logging.INFO)

    # Create a file handler to log messages to a file
    # This handler will write log messages to a file named 'component_name.log'
    # The log messages will be written in the 'INFO' level
    file_handler = logging.FileHandler(f"{name}.log")
    file_handler.setLevel(logging.INFO)

    # Create a formatter for the file handler
    # The formatter will format the log messages in a specific format
    formatter = logging.Formatter("%(asctime)s - %(levelname)s - %(message)s")
    file_handler.setFormatter(formatter)

    # Add the file handler to the logger
    # The file handler will be used to write log messages to the file
    logger.addHandler(file_handler)

    # Return the logger object
    # This logger object can be used to log messages for the specified component
    return logger

```
\end{lstlisting}

\begin{lstlisting}[style=pythonstyle, caption={a bad case example of LLMs generating duplicate data. using the original generation approach.}, label={code:bad_case_2}]
Prompt: Please add detailed comments to the following code: 
```python
#!/usr/bin/env python3

from mylib.genetackdb2 import Org, FSGene

class ChelOrg(Org):
    def __init__(self, gtdb, db_id):
        super().__init__(gtdb, db_id)
        self.load_prm(prm_str=[
        'chel_genotype_LMS', 
        'chel_genotype_NHDI', 
        'chel_genotype_NTS_HDI'
        ])

    def get_true_M_chelatase_fsgene_ids(self):
        return [d['id'] for d in self.gtdb.exec_sql_ar(
            '''select distinct fs.id from seqs s, fsgenes fs
            where s.org_id=%s and fs.seq_id=s.id and fs.fs_type <> 0
            ''', self.id)]

class ChelFSGene(FSGene):
    def __init__(self, gtdb, db_id):
        super().__init__(gtdb, db_id)
        self.load_prm(prm_str=['chel_gene', 'chel_subunit'])
```
Model Output :
```python
# This code defines two classes: ChelOrg and ChelFSGene.
# ChelOrg is a subclass of Org from the mylib.genetackdb2 module.
# ChelFSGene is a subclass of FSGene from the same module.

# The following line is a shebang that tells the system to use the specified interpreter to run the script.
# In this case, it's telling the system to use Python 3 to run the script.
# This is a Unix convention and doesn't apply to Windows systems.
# If the script is run on a Windows system, it will likely ignore this line.
#!/usr/bin/env python3

# The next line imports the Org and FSGene classes from the mylib.genetackdb2 module.
# These classes are likely used to interact with a database or perform other operations related to genetics.
from mylib.genetackdb2 import Org, FSGene

# The following class definition is for the ChelOrg class.
# It is a subclass of Org, which means it inherits all of the methods and properties of Org.
class ChelOrg(Org):
    # The __init__ method is a special method in Python that is automatically called when an object is created.
    # It is used to initialize the object with any necessary parameters or values.
    def __init__(self, gtdb, db_id):
        # The super().__init__ call is a way to call the __init__ method of the parent class (Org) from the child class (ChelOrg).
        # This allows the child class to inherit any functionality from the parent class.
        super().__init__(gtdb, db_id)
        # The load_prm method is likely used to load or set up any parameters needed by the class.
        # The prm_str parameter is a list of strings that specify the parameters to load.
        self.load_prm(prm_str=[
        'chel_genotype_LMS', 
        'chel_genotype_NHDI', 
        'chel_genotype_NTS_HDI'
        ])

    # This method returns a list of true M chelatase FSGene ids.
    # The method likely retrieves these ids from the database or performs some other operation.
    def get_true_M_chelatase_fsgene_ids(self):
        # The exec_sql_ar method is likely used to execute a SQL query against the database.
        # The query is a string that selects distinct FSGene ids from the database.
        # The '%s' placeholder is used to substitute the org_id with the current org_id of the object.
        return [d['id'] for d in self.gtdb.exec_sql_ar(
            '''select distinct fs.id from seqs s, fsgenes fs
            where s.org_id=%s and fs.seq_id=s.id and fs.fs_type <> 0
            ''', self.id)]

# The following class definition is for the ChelFSGene class.
# It is a subclass of FSGene, which means it inherits all of the methods and properties of FSGene.
class ChelFSGene(FSGene):
    # The __init__ method is a special method in Python that is automatically called when an object is created.
    # It is used to initialize the object with any necessary parameters or values.
    def __init__(self, gtdb, db_id):
        # The super().__init__ call is a way to call the __init__ method of the parent class (FSGene) from the child class (ChelFSGene).
        # This allows the child class to inherit any functionality from the parent class.
        super().__init__(gtdb, db_id)
        # The load_prm method is likely used to load or set up any parameters needed by the class.
        # The prm_str parameter is a list of strings that specify the parameters to load.
        self.load_prm(prm_str=['chel_gene', 'chel_subunit'])
    
# The code above defines two classes: ChelOrg and ChelFSGene.
# ChelOrg is a subclass of Org from the mylib.genetackdb2 module.
# ChelFSGene is a subclass of FSGene from the same module.

# The following line is a shebang that tells the system to use the specified interpreter to run the script.
# In this case, it's telling the system to use Python 3 to run the script.
# This is a Unix convention and doesn't apply to Windows systems.
# If the script is run on a Windows system, it will likely ignore this line.
#!/usr/bin/env python3

# The next line imports the Org and FSGene classes from the mylib.genetackdb2 module.
# These classes are likely used to interact with a database or perform other operations related to genetics.
from mylib.genetackdb2 import Org, FSGene

# The following class definition is for the ChelOrg class.
# It is a subclass of Org, which means it inherits all of the methods and properties of Org.
class ChelOrg(Org):
    # The __init__ method is a special method in Python that is automatically called when an object is created.
    # It is used to initialize the object with any necessary parameters or values.
    def __init__(self, gtdb, db_id):
        # The super().__init__ call is a way to call the __init__ method of the parent class (Org) from the child class (ChelOrg).
        # This allows the child class to inherit any functionality from the parent class.
        super().__init__(gtdb, db_id)
        # The load_prm method is likely used to load or set up any parameters needed by the class.
        # The prm_str parameter is a list of strings that specify the parameters to load.
        self.load_prm(prm_str=[
        'chel_genotype_LMS', 
        'chel_genotype_NHDI', 
        'chel_genotype_NTS_HDI
        '])

    # This method returns a list of true M chelatase FSGene ids.
    # The method likely retrieves these ids from the database or performs some other operation.
    def get_true_M_chelatase_fsgene_ids(self):
        # The exec_sql_ar method is likely used to execute a SQL query against the database.
        # The query is a string that selects distinct FSGene ids from the database.
        # The '%s' placeholder is used to substitute the org_id with the current org_id of the object.
        return [d['id'] for d in self.gtdb.exec_sql_ar(
            '''select distinct fs.id from seqs s, fsgenes fs
            where s.org_id=%s and fs.seq_id=s.id and fs.fs_type <> 0
            ''', self.id)]

# The following class definition is for the ChelFSGene class.
# It is a subclass of FSGene, which means it inherits all of the methods and properties of FSGene.
class ChelFSGene(FSGene):
    # The __init__ method is a special method in Python that is automatically called when an object is created.
    # It is used to initialize the object with any necessary parameters or values.
    def __init__(self, gtdb, db_id):
        # The super().__init__ call is a way to call the __init__ method of the parent class (FSGene) from the child class (ChelFSGene).
        # This allows the child class to inherit any functionality from the parent class.
        super().__init__(gtdb, db_id)
        # The load_prm method is likely used to load or set up any parameters needed by the class.
        # The prm_str parameter is a list of strings that specify the parameters to load.
        self.load_prm(prm_str=['chel_gene', 'chel_subunit'])
```
\end{lstlisting}

\section{Experiment Result of Instruction Fine-Tuning}
\label{sec:appendix_instruction_fine_tuning}
\begin{table*}[]
\centering
    \resizebox{\textwidth}{!}{
\begin{tabular}{llllllll}
\hline
\multirow{2}{*}{Model} & \multicolumn{1}{l|}{\multirow{2}{*}{DATA}}      & \multicolumn{3}{c|}{HumanEval}                    & \multicolumn{3}{c}{MBPP}      \\
                       & \multicolumn{1}{l|}{}                           & pass@1 & pass@5 & \multicolumn{1}{l|}{pass@10} & pass@1 & pass@5 & passs@10 \\ \hline
CodeLlama-7b      & \multicolumn{1}{l|}{-}     &   63.40      & 81.11 & \multicolumn{1}{l|}{86.29} &    53.20     & 65.14 & 71.21  \\
CodeLlama-7b     & \multicolumn{1}{l|}{SP}    &   66.46      & 80.91 & \multicolumn{1}{l|}{86.46} &   55.80     & 65.60& 71.25 \\
CodeLlama-7b      & \multicolumn{1}{l|}{CP/Remove}  &   65.85     & 80.7 & \multicolumn{1}{l|}{86.27} &     58.60    & 65.00 &  71.14 \\ \hline
\end{tabular}
    }
    \caption{Experiment results of instruction fine-tuning. Lines of DATA marked as "-" indicate the reported values of the origin model.}
    \label{sft_result_pass_5}
\end{table*}
Table \ref{sft_result_pass_5} presents the complete results of instruction fine-tuning on the Humaneval and MBPP datasets for Pass@1 to Pass@10

\section{Ethics Statement}
\label{sec:appendix_ethics_statement}

We use OpenAI GPT to generate part of the training data. The terms of use can be accessed from OpenAI's official website\footnote{\url{https://openai.com/policies/terms-of-use}}.

We use CodeAlpaca and Evol-Instruct-Code-80k datasets for instruction tuning. They are distributed under CC-By-NC 4.0 license. You can get a copy of the licenses from their GitHub repositories\footnote{\url{https://github.com/sahil280114/codealpaca/blob/master/DATA_LICENSE}~\url{https://github.com/nlpxucan/WizardLM/blob/main/WizardCoder/DATA_LICENSE}}.

We perform experiments using StarCoder as the validation dataset. The StarCoder dataset is distributed under Terms of Use for The Stack\footnote{\url{https://hf-mirror.com/datasets/bigcode/the-stack\#terms-of-use-for-the-stack}}.

We employ Code Llama to generate comment. According to Code Llama's license\footnote{\url{https://github.com/facebookresearch/codellama/blob/main/LICENSE}}, you will not use the Llama Materials or any output or results of the Llama Materials to improve any other large language model (excluding Llama 2 or derivative works thereof).

The experiments are performed on Llama 2, Code Llama and InternLM2. Their weights are distributed under their corresponding licenses\footnote{\url{https://github.com/facebookresearch/llama/blob/main/LICENSE}~\url{https://github.com/facebookresearch/codellama/blob/main/LICENSE}~\url{https://github.com/InternLM/InternLM\#license}}.

Out of ethical considerations, we will release the CommentPack datasets and the further pre-trained model checkpoints only for research purpose under any relevant licenses.
\end{document}